\documentclass[runningheads]{llncs}

\usepackage[numbers]{natbib}
\usepackage[preprint]{neurips_2024}

\usepackage[utf8]{inputenc} 
\usepackage[T1]{fontenc}    
\usepackage{url}            
\usepackage{booktabs}       
\usepackage{amsfonts}       
\usepackage{nicefrac}       
\usepackage{microtype}      
\usepackage{xcolor}    

\usepackage{subcaption}
\usepackage{graphicx}
\usepackage{booktabs}
\usepackage{comment}
\usepackage[accsupp]{axessibility}  

\usepackage[hidelinks]{hyperref}

\usepackage{orcidlink}

\usepackage{multirow}
\usepackage{float}
\usepackage{adjustbox}
\usepackage{makecell}

\begin{document}

\title{The Overfocusing Bias of Convolutional Neural Networks: A Saliency-Guided Regularization Approach}

\titlerunning{Abbreviated paper title}

\author{David Bertoin$\dagger$\\
  IRT Saint-Exupéry \\
  IMT, INSA Toulouse\\
  ANITI\\
  Toulouse, France\\
  \texttt{bertoin@insa-toulouse.fr} \\
\And
 Eduardo Hugo Sanchez$\dagger$ \\
  IRT Saint-Exupéry \\
  Toulouse, France\\
  \texttt{eduardo.sanchez@irt-saintexupery.com}
 \And 
  Mehdi Zouitine$\dagger$ \\
  IRT Saint-Exupéry \\
  IMT, Université Paul Sabatier\\
  Toulouse, France\\
  \texttt{mehdi.zouitine@irt-saintexupery.com}
  \And
  Emmanuel Rachelson\\
  ISAE-SUPAERO \\
  Université de Toulouse \\
  ANITI\\
  Toulouse, France\\
  \texttt{emmanuel.rachelson@isae-supaero.fr}
}

\maketitle

\def\thefootnote{$\dagger$}\footnotetext{These authors contributed equally to this work}\def\thefootnote{\arabic{footnote}}
\begin{abstract}
\label{sec:abstract}
Despite transformers being considered as the new standard in computer vision, convolutional neural networks (CNNs) still outperform them in low-data regimes. Nonetheless, CNNs often make decisions based on narrow, specific regions of input images, especially when training data is limited. This behavior can severely compromise the model's generalization capabilities, making it disproportionately dependent on certain features that might not represent the broader context of images.
While the conditions leading to this phenomenon remain elusive, the primary intent of this article is to shed light on this observed behavior of neural networks.
Our research endeavors to prioritize comprehensive insight and to outline an initial response to this phenomenon.
In line with this, we introduce Saliency Guided Dropout (SGDrop), a pioneering regularization approach tailored to address this specific issue.
SGDrop utilizes attribution methods on the feature map to identify and then reduce the influence of the most salient features during training. This process encourages the network to diversify its attention and not focus solely on specific standout areas. Our experiments across several visual classification benchmarks validate SGDrop's role in enhancing generalization. Significantly, models incorporating SGDrop display more expansive attributions and neural activity, offering a more comprehensive view of input images in contrast to their traditionally trained counterparts. 
\end{abstract}

\section{Introduction}
\label{sec:intro}


\begin{figure}[h!]
    \centering
    \begin{subfigure}[b]{0.6\linewidth}
        \centering
        \includegraphics[width=0.9\linewidth]{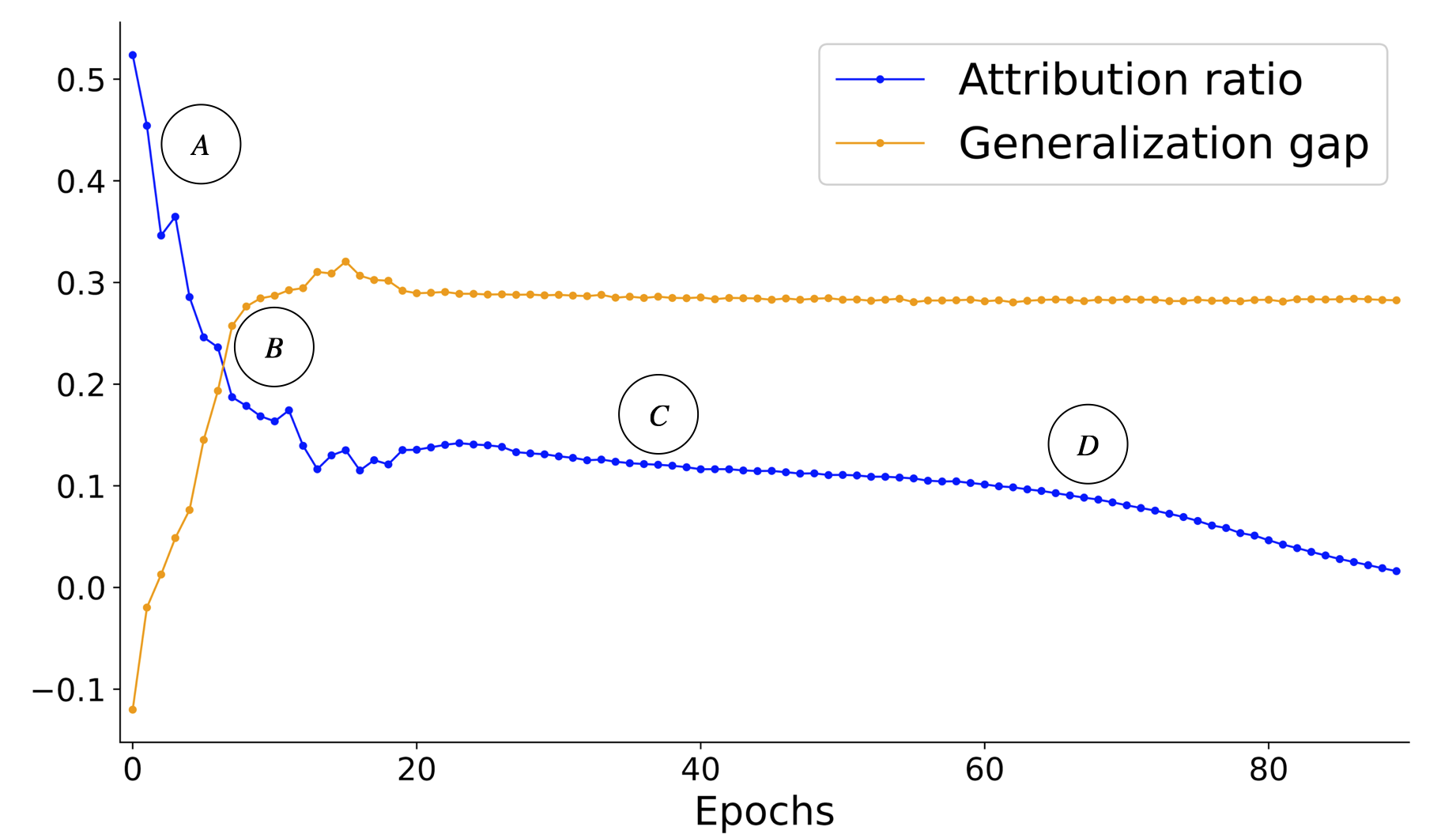} 
    \end{subfigure}
    \hfill 
    \begin{subfigure}[b]{0.39\linewidth}
        \centering
        \includegraphics[width=0.9\linewidth]{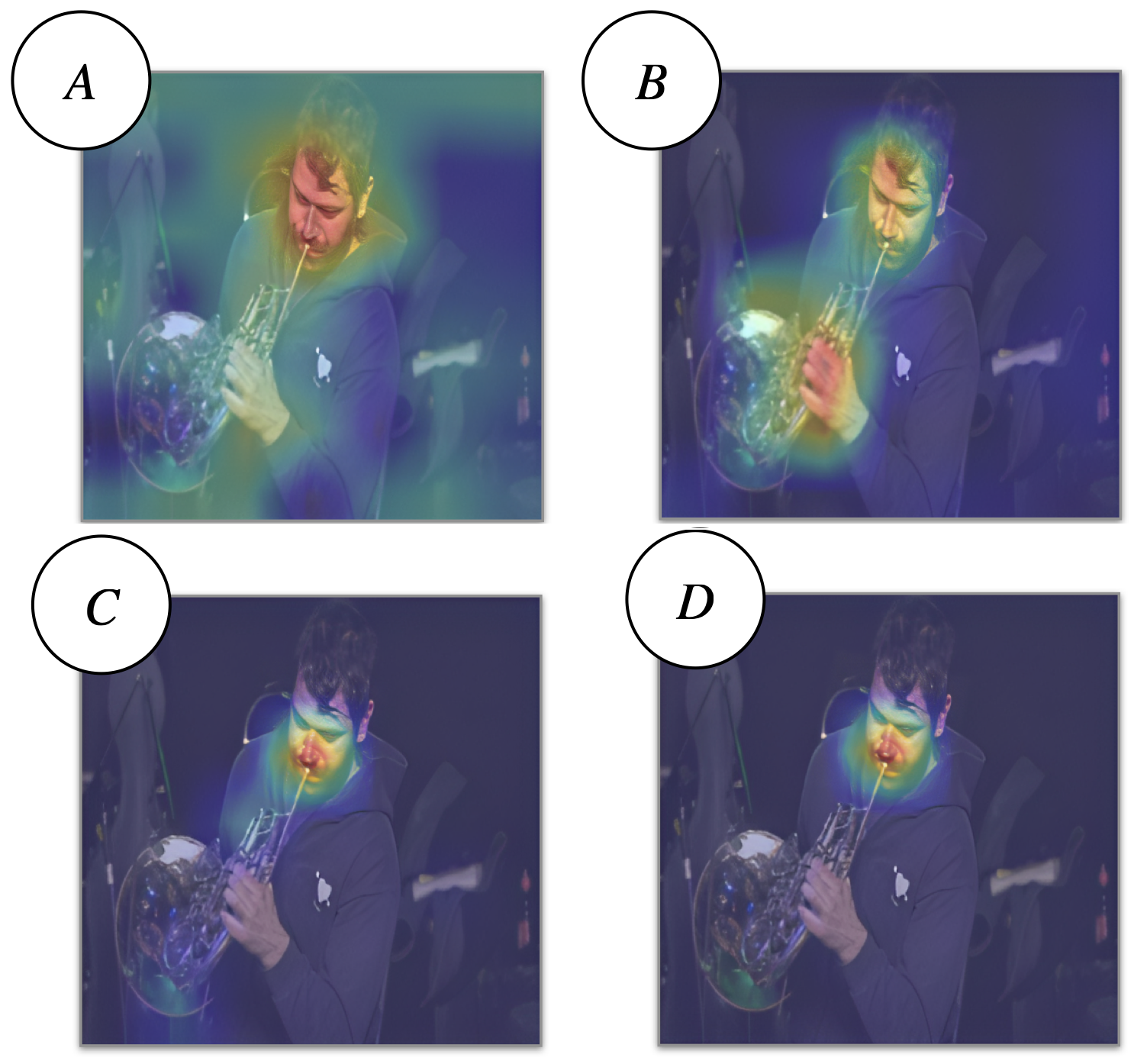} 
    \end{subfigure}
    \caption{Overfocusing phenomenon in neural networks: rapid concentration on small areas of images.}
    \label{fig:preliminary}
\end{figure}


Vision transformers \cite{dosovitskiy2020image} have emerged as the leading approach in various computer vision tasks, from image classification and object detection to instance segmentation.
However, when faced with limited datasets, convolutional neural networks \cite{krizhevsky2012imagenet, simonyan2015very, he2016deep,liu2022convnet} often demonstrate greater generalization potential, delivering more robust performance \cite{d2021convit, park2021vision, raghu2021vision, hassani2021escaping, liu2021efficient, lu2022bridging}. 
The growing performance of these networks over the last decade has been accompanied by a growing desire to understand the underlying mechanisms that drive their predictions, leading to the emergence of numerous interpretability methods.
Among the array of such tools, attribution methods \cite{simonyan2014deep, zeiler2014visualizing, sundararajan2017axiomatic, selvaraju2017grad, fong2017interpretable, petsiuk2018rise, fel2021look, novello2022making} stand out. They offer a visual gateway into the regions within images that neural networks deem pivotal for their decisions.
Such methods not only enhance our trust in these models but also provide an avenue for understanding potential pitfalls in their predictions.\\
\textbf{A preliminary experiment.} This work was inspired by a straightforward experiment.
We trained a VGG11 \cite{simonyan2015very} model on Imagenette \cite{imagenette_data}, a subset of the ImageNet \cite{deng2009imagenet} dataset, which encompasses 10,000 images spanning 10 classes derived from the original dataset.
During the network's training process, we continuously monitored the saliency maps generated using the Grad-CAM \cite{selvaraju2017grad} attribution method.
Figure \ref{fig:preliminary} chronicles the evolution of these saliency maps and the generalization gap between testing and training data throughout the training epochs\footnote{For an in-depth visual understanding, please refer to the accompanying videos available \href{https://youtu.be/WotjGpe88g8}{here}.}.
Our observations revealed an intriguing pattern. In the initial epochs, when the generalization gap was relatively small and the model's predictions were less accurate, the saliency maps broadly highlighted prominent objects, typically present in the foreground.
This behavior is intuitive, given that at this stage, neither the convolutional filters nor the fully connected layers have specialized to recognize distinct high-level patterns.
However, as training progressed, a shift occurred.
The saliency maps began to pinpoint singular elements associated with the predicted class.
By the latter epochs, these saliencies had narrowed significantly, focusing intently on minute details within the image.
Such pronounced saliency raises an intuitive concern: Is the network overfitting to its training data?
The shift from a broader focus on the entire object to an acute fixation on specific details hints at a model that might be leveraging minute training set nuances for predictions, potentially at the expense of broader generalization.\\
\textbf{Paper contributions.} Motivated by the observed behavior, we investigate the potential advantages that might emerge from promoting broader saliencies within neural networks.
To this end, we introduce Saliency-Guided Dropout (SGDrop), a regularization method that selectively drops the most salient features during training, steering the network towards a wider attentional focus on the input images.
Our empirical evaluations indicate that such an approach not only improves generalization across different data scales (including large datasets like ImageNet\cite{imagenet_data} and Places365 \cite{zhou2014learning}) but also leads to models showcasing expanded neural coverage and appearing more congruent with human perception.\\
\textbf{Paper layout.} 
Section \ref{sec:related_work} provides a review of recent literature, offering a critical 
perspective on our contribution and situating it in a broader research context.
Section \ref{sec:SGDrop} presents essential background information and introduces the notations used in this paper, followed by a detailed exposition of the development and implementation of the SGDrop regularization framework.
Section \ref{sec:experiments} conducts an empirical evaluation of SGDrop's effectiveness in broadening attribution scopes and its impact on model generalization. 
Section \ref{sec:limitations} addresses the limitations of our method, and Section \ref{sec:conclusion} provides a summary of our findings and concludes the paper.

\vspace{-0.25cm}

\section{Related work}
\label{sec:related_work}
\textbf{Attribution methods} have emerged as pivotal tools in the interpretability landscape, providing insights into the regions of input data that influence neural network decisions.
Broadly categorized, these methods fall into two main categories: \textit{white-box} and \textit{black-box}. 
White-box methods explain neural network decisions by directly examining and interpreting their internal parameters and structures.
Seminal work by Simonyan et al. \cite{simonyan2014deep} lays the foundation by computing attribution maps through the gradient of the output class relative to the input. Building upon this,  Smilkov et al.  \cite{smilkov2017smoothgrad} refines these maps, producing smoother saliencies by averaging gradients concerning perturbed inputs. Taking a more integrative approach, Sundararajan et al. \cite{sundararajan2017axiomatic} computes the integral of these gradients with respect to input features, tracing a path from a baseline to the input instance.
 Zhang et al. \cite{zhang2018top} introduce a probabilistic Winner-Take-All formulation for modeling the top-down attention and highlighting discriminative areas. Furthermore, Grad-CAM \cite{selvaraju2017grad}, identifies crucial input regions by harnessing the gradients of the target class channeled into the final convolutional layer.
Black-box methods interpret neural network decisions by analyzing model inputs and outputs without direct access to internal parameters or structures.
They typically involve computing statistics related to the impact of perturbing or occluding various regions of the input data on the model's predictions \cite{fong2017interpretable, petsiuk2018rise, fel2021look, novello2022making}.
Our SGDrop framework leverages attribution methods to regularize neural network training by selectively dropping the most salient pieces of information. Crucially, it is designed to be universally applicable and remains agnostic to the specific choice of attribution method, allowing for seamless integration with any interpretability approach.\\
\textbf{Saliency-Guided training} methods incorporate attention-derived insights to inform and regularize the learning process.
Ross et al. \cite{ross2017right} employ expert annotations to diminish the influence of irrelevant gradients.
Building on this concept, Ghaeini et al. \cite{ghaeini-etal-2019-saliency} and Fel et al.\cite{fel2022harmonizing}  strive to align network attributions with human explanations.
Gong et al. \cite{gong2021keepaugment} utilize saliency information to selectively enhance data augmentation on the less informative parts of the input. 
An array of studies \cite{walawalkar2020attentive, kim2020co, kim2020puzzle, uddin2020saliencymix, an2022saliency,kang2023guidedmixup} incorporate attribution methods to generate synthetic input/output pairs by merging critical regions from different inputs.
Yang et al. \cite{yang2022ad} intentionally discard certain high-attention areas to fine-tune the robustness of language models.
In self-supervised learning, Kakogeorgiou et al.\cite{kakogeorgiou2022hide} derive masks from a teacher's attention to design the masked image modeling objective.
Bertoin et al. \cite{bertoin2022look} introduce an auxiliary regularization task that predicts the attribution of an original input before augmentation, aiming to fortify resistance against distractions.
Ismail et al. \cite{ismail2021improving} apply a binary mask based on input attributions to preserve essential information, thus ensuring prediction consistency between original and masked inputs.
Approaches that more closely resemble our work include that of Choe and Shim\cite{choe2019attention}, who utilize channel-wise average pooling to create an attention map that alternately obscures pixels in feature maps with high and low intensities, prompting the network to rely on diverse regions for decision-making.
Similarly, Zunino et al. \cite{zunino2021excitation} employ excitation backpropagation to selectively deactivate neurons in feature maps based on their calculated saliency probabilities.
Distinctly, our SGDrop approach explicitly targets and removes the most influential neurons in the network's decision-making process in a deterministic manner.
This deliberate masking of the most critical information ensures that the network consistently focuses on alternative features in subsequent iterations. As we demonstrate in Section 5, this strategy enables SGDrop to achieve superior generalization while requiring the dropout of significantly fewer neurons compared to other methods.\\
\textbf{Preventing models from over-relying on specific features} serves as the foundational principle for a range of perturbation techniques.
Dropout methods \cite{srivastava2014dropout, ghiasi2018dropblock, tompson2015efficient} disrupt neuron co-adaptations by randomly omitting different neuron subsets during training.
 Pham and Le \cite{Pham2021autodropout} optimize these dropout patterns using a reinforcement learning controller, while Morerio et al. \cite{morerio2017curriculum} and do Santos et al. \cite{do2021maxdropout} suggest increasing dropout rates progressively or dropping the most active neurons, respectively.
 Bertoin and Rachelson \cite{bertoin2021local} disrupts input spatial correlations by locally permuting features in the latent space.
These methods are complemented at the pixel level by  DeVries and Taylor \cite{devries2017improved} and Yun et al. \cite{yun2019cutmix}, who propose removing or replacing image patches to mitigate model overfitting to particular input features.
These resonate with our method's goal of countering the network's tendency to overfocus on specific attributes and promoting a more uniform distribution of attributions across the input space.
While these approaches share our objective of preventing model over-reliance on a narrow set of features, our SGDrop technique distinctively leverages attribution methods  to precisely pinpoint areas for intervention.
By targeting specific regions for intervention, SGDrop allows the unaffected portions of the network to adapt effectively, which in turn ensures a more balanced focus across the entire feature spectrum.

\section{Saliency-Guided Dropout}
\label{sec:SGDrop}
\begin{figure*}[ht]
    \centering
    \includegraphics[width=0.8\linewidth]{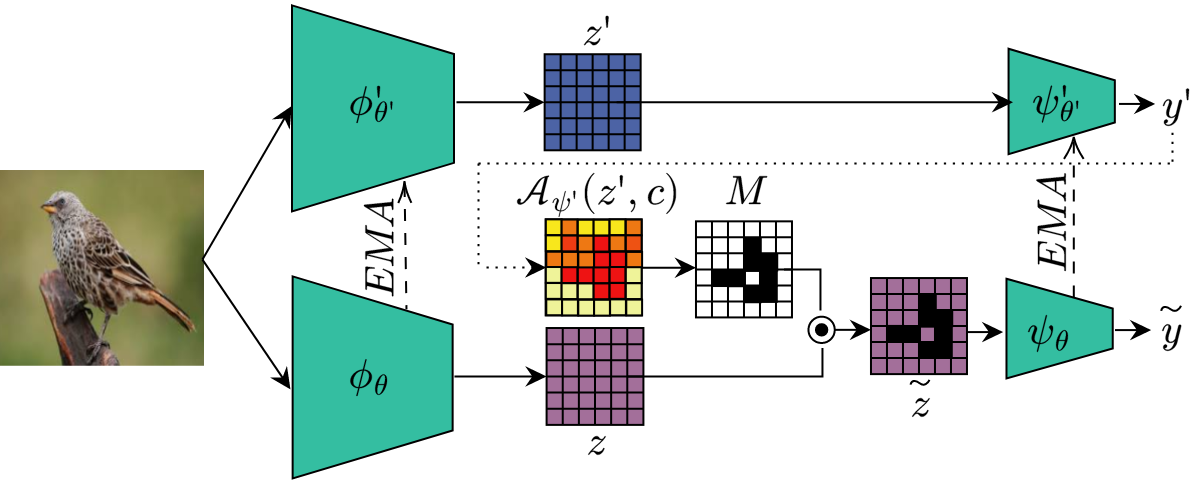}
    \caption{Overview of our proposed method SGDrop.}
    \label{fig:SGDrop}
\end{figure*}
We consider the general supervised classification problem, given by a distribution $p(x,c)$, where $x\in\mathcal{X}$ are inputs belonging to a finite-dimensional feature space, and $c\in \mathcal{C}$ are discrete labels taken from the finite set of classes $\mathcal{C}$.
Note that we set ourselves in a fairly general framework, where an element $x\in \mathcal{X}$ can be an image, but also a map of pre-computed features on a image, or even a plain vector of scalar values. 
We will assume however that the feature space $\mathcal{X}$ is $d$-dimensional, and hence isomorphic to $\mathbb{R}^d$.
With a slight abuse of notation, we shall write $\mathcal{Y} = \mathbb{R}^{|C|}$ the set of discrete probability distributions on $\mathcal{C}$ and will note $y_c$ the probability vector in $\mathbb{R}^{|Y|}$ which sets all probability mass on the $c$th item. 
A solution to the classification problem takes the form of a function $f$ within a hypothesis space $\mathcal{H}$, mapping $x\in \mathcal{X}$ to a distribution $f(x)\in \mathcal{Y}$ on classes. 
For notation convenience, we write $f_c(x)$ the $c$th component of $f(x)$, i.e. the probability estimate of class $c$.
Given a loss function $\ell(\hat{y},y)$ quantifying the cost of predicting distribution $\hat{y}\in \mathcal{Y}$ when the ground truth was $y\in \mathcal{Y}$, an approximate solution to the classification problem can be found by minimizing the empirical risk $\frac{1}{n} \sum_{i=1}^{n} \ell(f(x_i),y_{c_i})$, relying on a finite sample of input-label pairs ${(x_i,c_i)}_{i=1}^n$ drawn independently from $p$ \cite{vapnik1991principles}. 
Classically, $\ell$ is the cross-entropy between distributions $y_c$ and $f(x)$, and networks $f\in\mathcal{H}$ feature a softmax output layer to ensure their output is a discrete probability distribution. 
Throughout the remainder of this paper, we define $\mathcal{H}$ as the space of neural network functions with parameters $\theta$. 
We define the attribution (or saliency) map for class $c$, of a network $f \in \mathcal{H}$ evaluated in $x\in\mathcal{X}$, as an item $\mathcal{A}_f(x,c) \in \mathbb{R}^d$ that quantifies the contribution of each feature in $\mathcal{X}$ to the prediction probability for class $c \in \mathcal{C}$.
The method we propose is entirely agnostic of the way $\mathcal{A}_f(x,c)$ is computed, we simply assume some specific, adequate method provides such a measure of feature influence.

\subsection{Generic algorithm}

Our intention is to let a neural network learn a function which minimizes the empirical risk, while preventing the over-fitting due to  excessive reliance on specific features. 
Dropout \cite{srivastava2014dropout} aims to achieve the same goal by randomly masking neurons during the gradient descent process, so as to avoid over-specialization on minute features, and dispatch the network feature reliance on all relevant features. 
But the dropout process is myopic in the sense that, to avoid over specialization on a few neurons, it drops out a large population of neurons (in the order of 20 to 70 percent of a layer's neurons). 
By doing so, it prevents over-specialization on some features, but at the same time also drops out under-represented important features with high probability. 
The rationale (and expected strength) of the method we propose is that it is possible to identify a very small population of key neurons to drop out, while preserving many degrees of freedom in gradient descent. 
We posit attribution maps constitute a good proxy to pinpoint these neurons and hence introduce a generic method of Saliency-Guided Dropout (SGDrop).
Given a neural network $f$ with parameters $\theta$, SGDrop mitigates the network's over-reliance on a subset of features by redistributing its focus across neurons. 
For $\rho \in [0,1]$, let $q_\rho$ be the upper $\rho$-quantile of attribution map $\mathcal{A}_f(x,c)$, i.e. the attribution value below which one finds a proportion $1-\rho$ of values in $\mathcal{A}_f(x,c)$. 
The $q_\rho$-saliency dropout mask is then the vector $M_f(x,c,\rho) \in \{0, 1\}^d$ which has value 1 for features taking their values below the upper $\rho$-quantile, and 0 otherwise. 
That is $M_f(x,c,\rho)=\mathbb{I}[\mathcal{A}_f(x,c) \leq q_\rho]$, where $\mathbb{I}[\cdot]$ is the indicator function, assigning 1 if the attribution is below $q_\rho$, and 0 otherwise. 
We call \emph{saliency-regularized features} the map $\tilde{x} = x \odot M_f(x,c,\rho)$, where $\odot$ is the Hadamard product. 
Note that this regularized feature map depends on the actual label $c$ of input $x$, and corresponds to the original input $x$ where the identified most salient features for its associated class $c$ have been dropped out and set to zero. 
The empirical risk under saliency-guided dropout (SGDrop) is then $\frac{1}{n} \sum_{i=1}^{n} \ell(f(\tilde{x}_i),y_{c_i})$. 
In particular, the cross-entropy loss under saliency-guided dropout for an input $x$ having label $c$ is written:
\begin{equation*}
\ell(f(\tilde{x}),y_c) = - \log( f_c(x \odot M_f(x,c,\rho)) ).
\end{equation*}

\subsection{Practical implementation}

\textbf{Attribution in the latent space.}
We consider a network $f = \psi\circ \phi$ composed of an encoder $\phi$ and a classifier $\psi$. 
In implementing our SGDrop method, we chose to compute attributions on the latent features $z$, produced by the encoder segment $\phi$ of the network.
We favor this approach for its computational efficiency, as attributions on latent features are faster to compute than on input pixels.
Additionally, it results in attributions that are inherently smoother and more interpretable, capturing high-level concepts that map to extensive areas in the pixel space, thereby focusing on semantically rich and coherent structures within the data.
Once the latent features $ z $ are extracted by the encoder $ \phi $ of the network, they are fed into the classifier part $ \psi $, and attribution map $\mathcal{A}_\psi(z, c)$ directs our SGDrop approach.\\
\textbf{Attribution method.}
To delineate feature influences accurately, we implement a straightforward yet effective attribution method. It harnesses the gradients of the class scores relative to the activations, similar in spirit to the principles of Grad-CAM, yet distinct in that it maintains the full detail of the feature maps by eschewing channel averaging. Attribution is thus computed by element-wise multiplication of the positive gradient components and feature activations:
\begin{equation*}
\mathcal{A}_\psi(z,c) = \text{ReLU}\left(\nabla_z \psi_c(z) \odot z\right),
\end{equation*}
where $\nabla_z \psi_c(z)$ is the gradient of the class score with respect to the feature map $z$, and $\text{ReLU}$ ensures that, as for Grad-CAM, only gradients with the same sign as the activation are accounted for.\\
\textbf{To stabilize training} and facilitate a practical implementation, we compute the attribution map using an exponential moving average (EMA) $\theta'$ of the network parameters $\theta$. 
Parameters $\theta'$ define a network $f' = \psi' \circ \phi'$ and we use the saliency map $\mathcal{A}_{\psi'}(\phi'(x),c)$ in place of $\mathcal{A}_{\psi}(z,c)$ to define the $q_\rho$-saliency dropout mask and, in turn, the saliency-regularized features $\tilde{z}$. 
Figure \ref{fig:SGDrop} graphically illustrates the foundational principles of our implementation of the SGDrop method.



\section{Experiments}
\label{sec:experiments}
In this section, we present a comprehensive empirical evaluation of SGDrop, focusing on its efficacy in broadening attributions and the consequential improvements in network interpretability, neuron coverage, and generalization capabilities. 
We also delve into the critical design choices underpinning SGDrop, shedding light on their impact and the underlying mechanisms driving these improvements.  
Hyperparameters, network architectures, and implementation choices are summarized in Appendix A. 
One salient feature of SGDrop we wish to highlight from the start is the small value for $\rho$. By default, $\rho=0.01$, which corresponds to dropping 250 neurons out of $25,000$ within the VGG16 feature maps. In contrast, a standard value for dropout probabilities is within the $[0.2,0.7]$ range. The discussion below should be taken with these values in mind.


\begin{figure}[h!]
    \centering
    \begin{subfigure}[b]{0.49\linewidth}
        \centering
        \includegraphics[width=\linewidth]{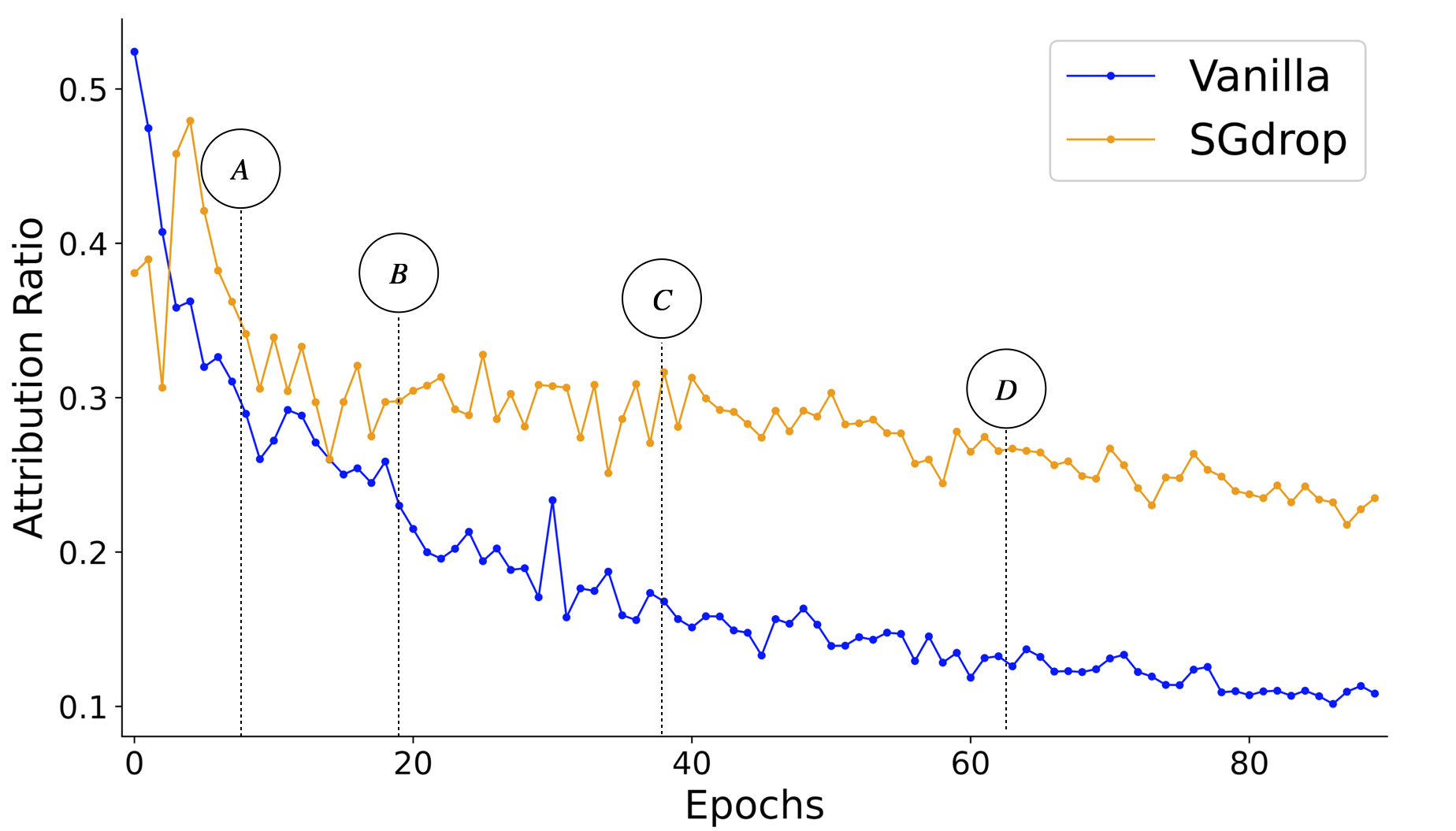} 
        \caption{Attribution Ratio Comparison}
        \label{fig:attribution_ratio}
    \end{subfigure}
    \hfill 
    \begin{subfigure}[b]{0.49\linewidth}
        \centering
        \includegraphics[width=\linewidth]{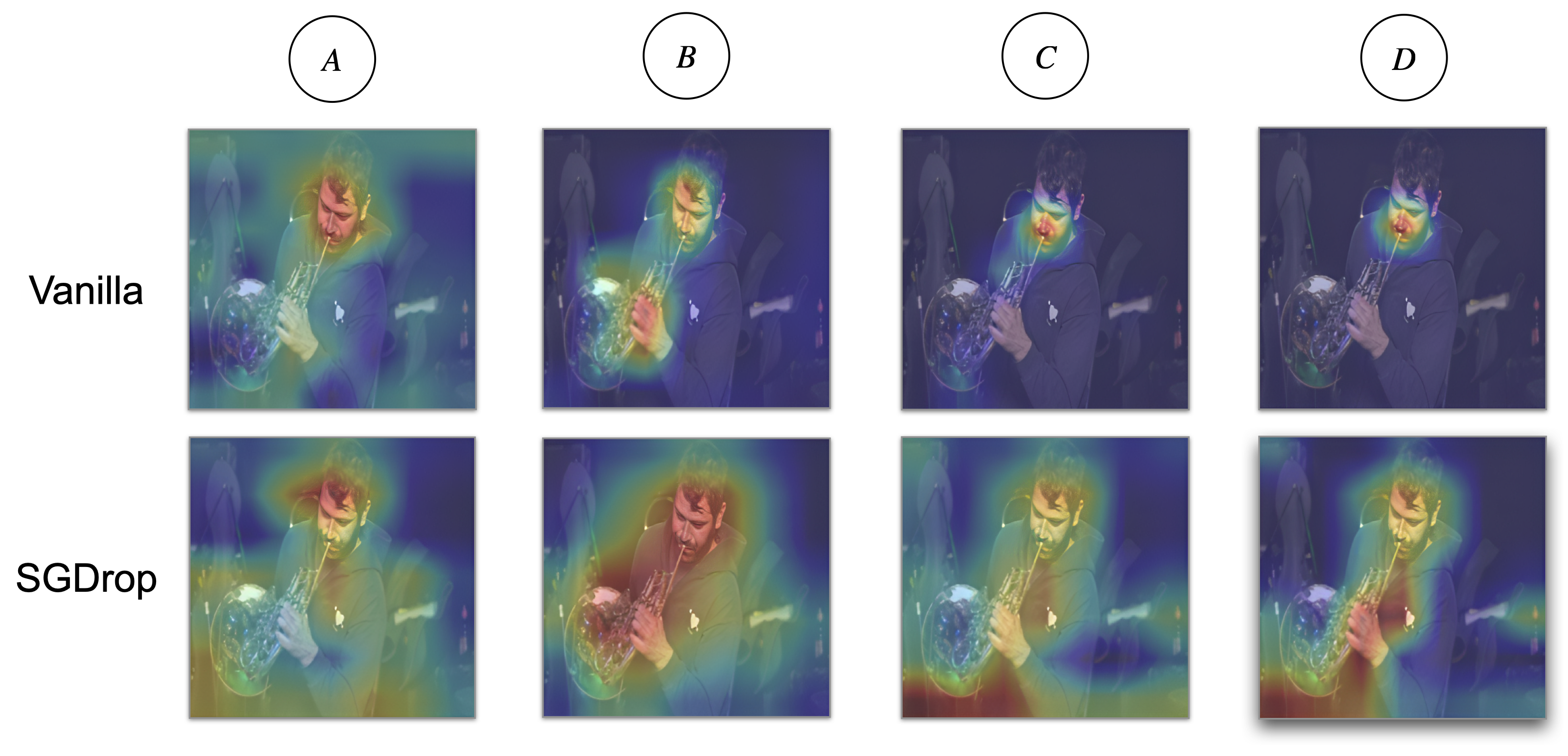} 
        \caption{Saliency Comparison}
        \label{fig:saliency_comp}
    \end{subfigure}
    \caption{Variation in attribution area ratios and associated saliency maps during training.}
    \label{fig:preliminaryv2}
\end{figure}

\subsection{Back to the preliminary experiment}

We employed SGDrop in conjunction with a VGG11 \cite{simonyan2015very} architecture to investigate its effectiveness in counteracting the attribution shrinkage phenomenon, as initially observed in our preliminary experiments (Section \ref{sec:intro}).
Throughout the training process, we meticulously quantified the attribution map ratio on the input images across the entire test set after each epoch.
This ratio was calculated as the proportion of pixels in the attribution map exceeding a threshold of 0.5 (with the maximum value normalized to 1), effectively capturing the extent of significant feature activations. 
As depicted in Figure \ref{fig:preliminaryv2},\footnote{For an in-depth visual understanding, please refer to the accompanying videos in the supplementary material.} the network trained with SGDrop consistently sustains a broader attribution scope over successive epochs, demonstrating its potential to counteract the narrowing of focus typically observed in conventional training regimes.
\\

\subsection{Saliencies and neural coverage}
\begin{figure}[ht]
    \begin{minipage}[c]{0.5\textwidth} 
        \centering
        \begin{subfigure}[b]{0.3\textwidth}
            \centering
                \includegraphics[height=1.6cm]{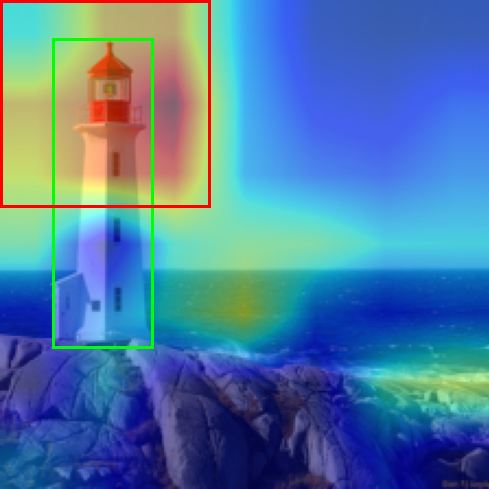}
        \end{subfigure}%
        \begin{subfigure}[b]{0.3\textwidth}
            \centering
                \includegraphics[height=1.6cm]{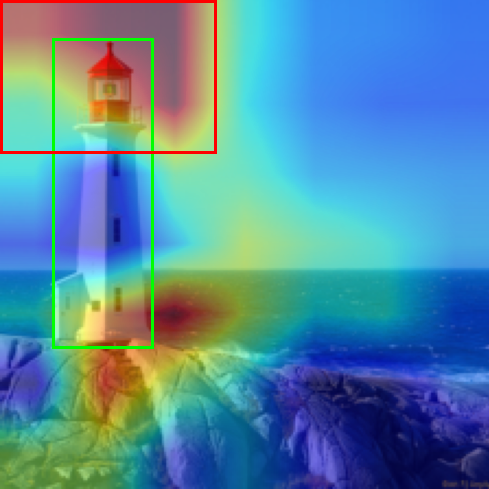}
        \end{subfigure}
        \begin{subfigure}[b]{0.3\textwidth}
            \centering
                \includegraphics[height=1.6cm]{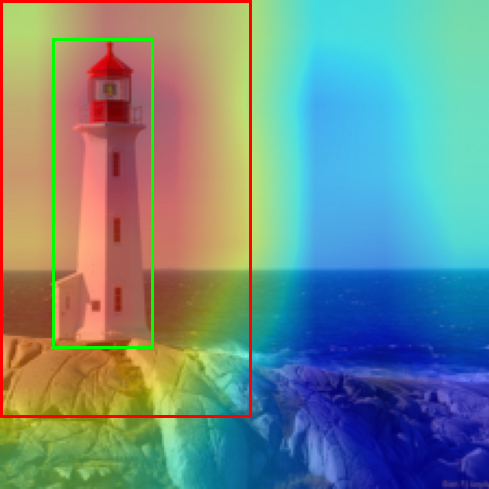}
        \end{subfigure}
        \begin{subfigure}[b]{0.3\textwidth}
            \centering
                \includegraphics[height=1.6cm]{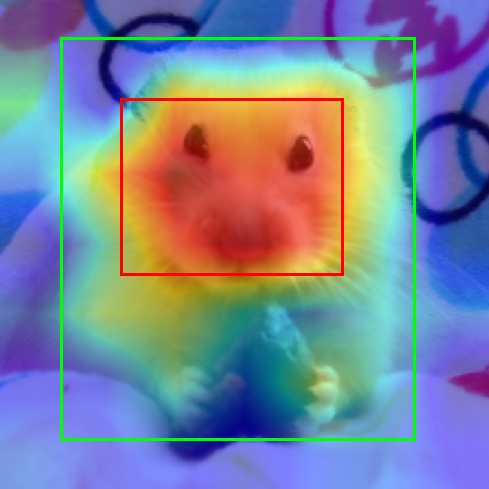}
        \end{subfigure}%
        \begin{subfigure}[b]{0.3\textwidth}
            \centering
                \includegraphics[height=1.6cm]{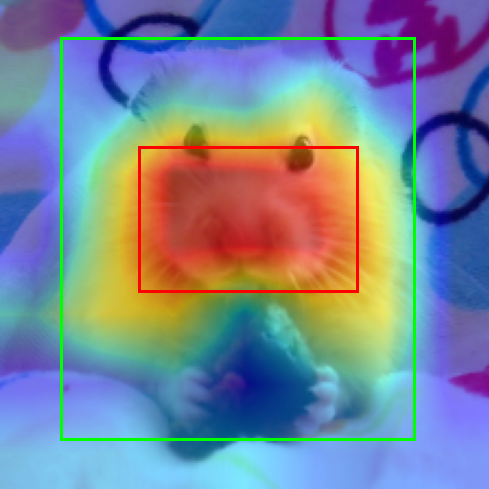}
        \end{subfigure}
        \begin{subfigure}[b]{0.3\textwidth}
            \centering
                \includegraphics[height=1.6cm]{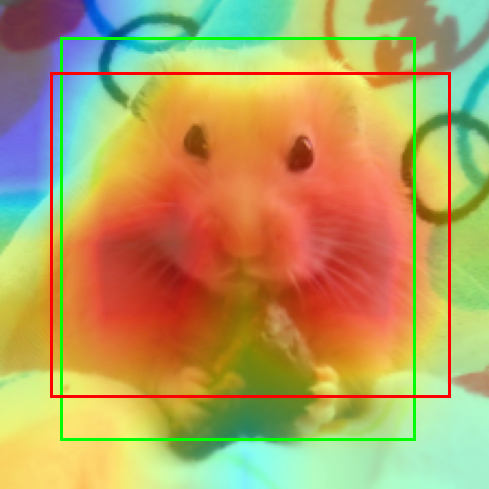}
        \end{subfigure}
        \\
        \begin{subfigure}[b]{0.3\textwidth}
            \centering
                \includegraphics[height=1.6cm]{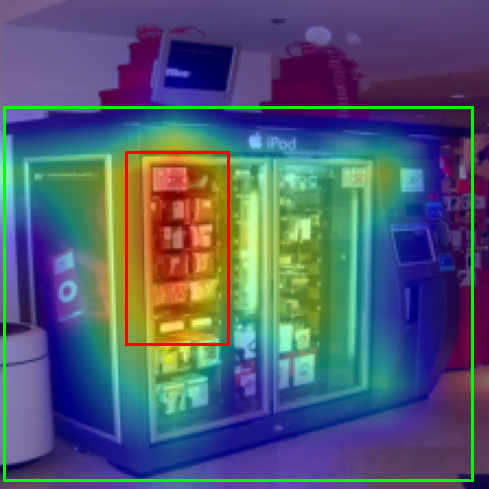}
            \caption{Vanilla}
        \end{subfigure}%
        \begin{subfigure}[b]{0.3\textwidth}
            \centering
                \includegraphics[height=1.6cm]{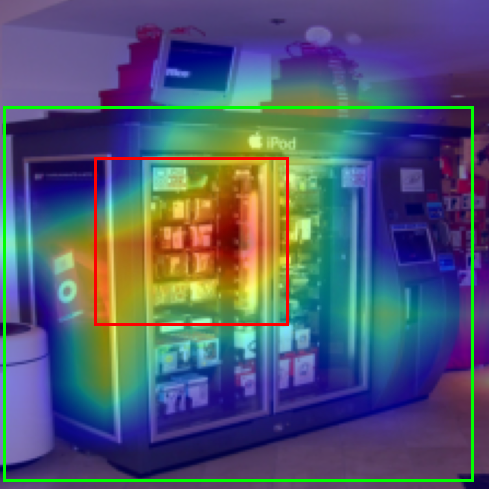}
            \caption{Dropout}
        \end{subfigure}
        \begin{subfigure}[b]{0.3\textwidth}
            \centering
                \includegraphics[height=1.6cm]{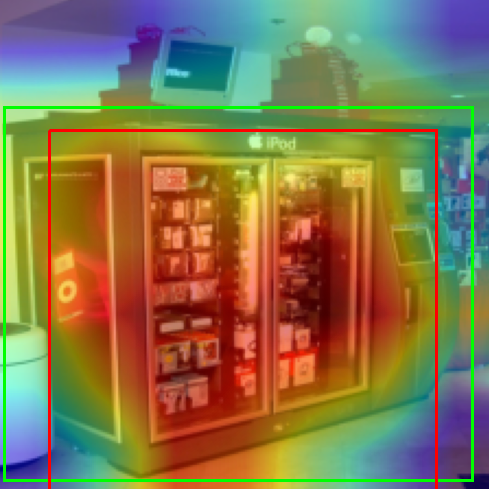}
            \caption{SGDrop}
        \end{subfigure}
        \caption{Qualitative comparison of saliency map for vanilla, dropout, and SGDrop training.
        Ground truth bounding boxes in green and attribution-derived bounding boxes in red.}
        \label{fig:rho}
    \end{minipage}%
    \hfill
    \begin{minipage}[c]{0.45\textwidth}
    \captionof{table}{Attribution quality metrics on the ImageNet test set.}
    \centering
    \begin{tabular}{@{}lccc}
    \toprule
      Method  & \makecell{Area \\ ratio} & \makecell{Hit \\ ratio} & \makecell{Harmonization \\ score}\\
    \midrule
        Vanilla & 0.01 & 0.32 & 0.23\\
        Dropout & 0.02 & 0.39 & 0.26\\
        SGDrop & \textbf{0.14} & \textbf{0.83} & \textbf{0.33}\\
    \bottomrule
    \end{tabular}
    \label{tab:metrics}
    \captionof{table}{Neuron coverage on the entire network and the feature map.}
    \begin{tabular}{@{}lcc}
    \toprule
      Method & \makecell{Global \\ coverage} & \makecell{Feature map \\ coverage} \\
    \midrule
        Vanilla & 0.40 & 0.09 \\
        Dropout & 0.38 & 0.13 \\
        SGDrop & 0.39 & 0.58 \\
    \bottomrule
    \end{tabular}
        
        \label{tab:neuron_coverage}
    \end{minipage}
\end{figure}

We rigorously quantify SGDrop's impact on attribution maps, using the testing set of the large-scale ImageNet \cite{deng2009imagenet} dataset. 
We trained a VGG16 network over 90 epochs using three configurations: SGDrop, classical dropout \cite{srivastava2014dropout}, and vanilla (no regularization).
Our analysis is structured around three pivotal metrics: the attributions spread, their concordance with actual labels, and their alignment with human interpretations (Table \ref{tab:metrics}).\\
\textbf{The attribution spread} metric, akin to our preliminary experiment, is derived by measuring the average area of the attribution maps where pixel values exceed a threshold of 0.5. 
SGDrop achieves an average attribution area ratio of 0.14. 
This is significantly broader than the baseline models: about tenfold that of vanilla models (0.01), and classical dropout (0.02). These findings demonstrate SGDrop's effectiveness in generating attribution maps that are considerably more extensive and inclusive than those produced by conventional methods.\\
\textbf{The adequacy of the attributions with real labels} metric is calculated by generating bounding boxes from attribution values at or above 0.5 and considering a ``hit'' when their intersection-over-union (IoU) with ground truth boxes is no less than 0.5.
SGDrop demonstrates remarkable efficacy here, achieving a hit score of 0.83, which significantly overshadows the scores of 0.39 for dropout and 0.32 for vanilla models, as visually depicted in Figure \ref{fig:rho}.\\
\textbf{The harmonization score} metric introduced by Fel et al. \cite{fel2022harmonizing}, inspects the extent to which network attributions align with human-generated explanations. Here again, SGDrop notably excels, validating its capacity to produce saliency maps that resonate more closely with human perceptual and interpretative patterns.
This holistic analysis underscores the comprehensive benefits of SGDrop, not only in enhancing attribution breadth but also in improving the overall interpretability and alignment with human cognitive patterns.\\
\textbf{Neuron coverage.} 
In addition to the attribution-focused metrics, following the principles of Pei et al. \cite{pei2017deepxplore}, we measure the neuron coverage across the three network configurations, on 10 random batches of 64 samples from the ImageNet test set (Table \ref{tab:neuron_coverage}).
Neuron coverage is defined as the proportion of activated neurons in a neural network, computed by assessing whether the activation of each neuron exceeds a specified threshold during the inference of a set of inputs. In line with the ReLU activation function, we set our neuron activation threshold at 0, identifying neurons that are activated and contribute to the network's decision process.
While the global neuron coverage across all layers remains relatively consistent across the three models, a notable distinction emerges upon closer inspection of the last convolutional layer. 
There, SGDrop demonstrates a remarkable enhancement, exhibiting up to 4.5 times more activated neurons compared to its counterparts.
This significant increase in neuron coverage within the critical feature-extraction layer of SGDrop underscores its proficiency in mobilizing a more diverse set of neurons for decision-making.
This expanded activation pattern may play a crucial role in the enhancements observed in both attribution quality and the interpretability of the model, as evidenced by our previous metrics.




\subsection{Generalization performance} 
We evaluate whether SGDrop's broadened attributions positively impact generalization across various low data regimes datasets.
These include CIFAR-10 \cite{cifar10_data} and STL10 \cite{coates2011analysis}, two low-resolution datasets with 60,000 and 5,000 training images respectively, both across 10 classes; Imagenette and Imagewoof \cite{imagenette_data}, subsets of ImageNet \cite{imagenet_data} containing approximately 10,000 images in 10 classes; and three remote sensing datasets for scene classification, namely RSICD \cite{rsicd_data}, RESISC45 \cite{resisc45_data}, and PatternNet \cite{patternnet_data}, encompassing 10,921, 31,500, and 30,400 images across 31, 45, and 38 classes, respectively.
We conduct a comparative analysis, benchmarking our approach against models employing standard Dropout \cite{srivastava2014dropout} and Excitation Dropout \cite{zunino2021excitation}, both with a dropout probability of $p=0.5$ and models trained with concurrent saliency-guided methods like the one from Ismail et al. \cite{ismail2021improving} and SGQN \cite{bertoin2022look}. In this context, we train a VGG16, a Resnet50 \cite{he2016deep} and, a ConvNeXt \cite{liu2022convnet} for 80 epochs, applying minimal data augmentation which involves only resizing images to 224x224 pixels. An exception is made for CIFAR10, where we utilize a smaller convolutional neural network that aligns with the one used in \cite{zunino2021excitation}, and for STL10, where a VGG9 is employed. 
As summarized in Tables \ref{tab:Comparison_with_methods_across_datasets} and \ref{tab:cifar_stl10}, the results, averaged over three runs, consistently show SGDrop outperforming all baselines.
On datasets with smaller images, SGDrop performs comparably to other baselines on CIFAR10, while significantly outperforming them on STL10, achieving an improvement of over 7\% compared to Excitation Dropout.
On the VGG16 architecture in particular, SGDrop enables remarkable improvements, registering more than 18\% increase on Imagenette, 35\% on Imagewoof, 30\% on RSICD, 24\% on RESISC45, and 4\% on PatternNet compared to the next best performing method.
The ResNet50 architecture also benefits from SGDrop, albeit to a slightly lesser extent, showing improvements of more than 3\% on Imagenette, 12\% on Imagewoof, 2\% on RSICD, 6\% on RESISC45, and marginal gains on PatternNet.
Interestingly, VGG16 models trained with SGDrop consistently outperform ResNet50 models across all datasets, a notable observation considering ResNets typically surpass VGGs in baseline comparisons.
This underlines the exceptional efficacy of SGDrop, particularly in enhancing VGG architectures.
SGDrop yields significant enhancements in performance for ConvNeXt, with improvements exceeding 5\% on Imagenette, 6\% on Imagewoof, and 2\% on RESISC45. It also provides slight improvements on RSICD and PatternNet, thereby showcasing its positive impact on modern architectures.
In order to further explore SGDrop's potential in large dataset scenarios, we apply it to the ImageNet and Places 365 \cite{zhou2014learning} datasets, using the official Pytorch \cite{paszke2019pytorch} ImageNet training script for both VGG16 and Resnet50, with and without data augmentation (DA) comprising horizontal flipping and random cropping.
As detailed in Table \ref{tab:large_data}, SGDrop markedly enhances VGG16's generalization performance across both datasets, irrespective of DA.
However, we observe no significant improvement with the Resnet50 architecture in these settings.
Notably, on Places365, VGG16's performance again rivals that of ResNet50.

\subsection{Transfer learning}
We assess the transferability of representations learned through SGDrop by conducting a transfer learning experiment. We train a VGG16 architecture on RESISC45 and subsequently fine-tune it on RSICD by keeping the feature extractor frozen (that is, training only the classification layer). Results, shown in Table \ref{tab:transfer_learning_exps_on_rsicd}, indicate that model utilizing SGDrop's features surpasses others, achieving over a 5\% improvement in performance.
SGDrop's design prevents neural networks from overfitting to confined image regions, thereby ensuring that the learned representations contain broader and more meaningful information, which enhances their applicability across datasets with similar attributes.

\begin{table*}[!ht]
  \centering
  \caption{Accuracy on the testing sets.}
  \begin{adjustbox}{width=\columnwidth,center}
  \begin{tabular}{@{}llccccc@{}}
    \toprule
    Network & Method             & Imagenette       & Imagewoof   & RSICD            & RESISC45         & PatternNet       \\
    \midrule
    \multirow{4}{3em}{VGG16} & Vanilla \cite{simonyan2015very}  & 67.21 $\pm$ 0.29 &  43.65 $\pm$ 0.67      & 55.60 $\pm$ 2.13 & 59.32 $\pm$ 2.71 & 90.35 $\pm$ 1.37  \\
                             & Dropout \cite{srivastava2014dropout}   & 68.66 $\pm$ 0.54 &  38.88 $\pm$ 0.92     & 58.40 $\pm$ 1.42 & 68.13 $\pm$ 0.90 & 94.61 $\pm$ 0.22  \\
                             & Excitation Dropout \cite{zunino2021excitation} & 68.48 $\pm$ 0.33 &  39.00 $\pm$ 0.37          & 56.33 $\pm$ 0.17 & 66.45 $\pm$ 1.40 & 93.64 $\pm$ 0.30  \\
                             & Ismail et al. \cite{ismail2021improving}   & 62.74 $\pm$ 1.01 &  38.02 $\pm$ 1.12       & 54.53 $\pm$ 1.46 & 67.49 $\pm$ 0.45 & 93.74 $\pm$ 1.12\\
                             & SGQN  \cite{bertoin2022look}   & 65.97 $\pm$ 2.21 &  38.14 $\pm$ 3.87       & 58.28 $\pm$ 0.09 & 70.82 $\pm$ 2.73 & 94.42 $\pm$ 0.23\\
    & SGDrop (ours)  & \textbf{81.00 $\pm$ 0.44} & \textbf{59.56 $\pm$ 0.11}      & \textbf{75.94 $\pm$ 2.28} & \textbf{84.60 $\pm$ 0.41} & \textbf{98.53 $\pm$ 0.14} \\
    \midrule
    
    \multirow{4}{3em}{ResNet50} & Vanilla \cite{he2016deep} & 74.96 $\pm$ 0.46 &  45.76 $\pm$ 0.13      & 69.35 $\pm$ 1.43 & 78.28 $\pm$ 0.48 & 98.52 $\pm$ 0.21  \\
                                & Dropout \cite{srivastava2014dropout}  & 74.08 $\pm$ 0.34 &  45.86 $\pm$ 0.08     & 67.64 $\pm$ 1.23 & 77.87 $\pm$ 0.29 & 98.56 $\pm$ 0.02 \\
                                & Excitation Dropout \cite{zunino2021excitation}  & 74.10 $\pm$ 0.42 &  48.16 $\pm$ 0.21       & 70.02 $\pm$ 0.17 & 78.34 $\pm$ 0.67 & 98.38 $\pm$ 0.14 \\
                                & Ismail et al. \cite{ismail2021improving}  & 72.58 $\pm$ 0.91 &  45.63 $\pm$ 0.91       & \textbf{71.78 $\pm$ 1.15} & 83.65 $\pm$ 0.03 & 98.42 $\pm$ 0.11 \\
                                & SGQN  \cite{bertoin2022look}   & 72.94 $\pm$ 1.66 &  45.58 $\pm$ 1.66       & 71.63 $\pm$ 0.37 & \textbf{84.40 $\pm$ 0.05} & 98.45 $\pm$ 0.05 \\
    & SGDrop (ours)  & \textbf{77.54 $\pm$ 0.45} & \textbf{53.63 $\pm$  0.60}      & 71.06 $\pm$ 1.11 & 82.75 $\pm$ 0.35 & \textbf{98.58 $\pm$ 0.09} \\
    \midrule
    
    \multirow{4}{3em}{ConvNeXt} & Vanilla  \cite{liu2022convnet}           &55.08 $\pm$ 2.01          &  24.64 $\pm$ 0.49           & 49.04 $\pm$ 1.01          & 67.89 $\pm$ 0.88          & 95.96 $\pm$ 0.27  \\
                                & Dropout \cite{srivastava2014dropout}             & 55.36 $\pm$ 2.7          &  24.64 $\pm$ 0.48          & 55.08 $\pm$ 1.10          & 71.68 $\pm$ 0.41          & 96.98 $\pm$ 0.00  \\
                                & Excitation Dropout \cite{zunino2021excitation}  & 62.47 $\pm$ 1.06          &  26.91 $\pm$ 0.29           & 54.94 $\pm$ 1.60          & 73.22 $\pm$ 1.23          & 96.90 $\pm$ 0.25  \\
                                & SGDrop  (ours)             & \textbf{67.42 $\pm$  2.12} & \textbf{32.79 $\pm$  0.61}  & \textbf{55.76 $\pm$ 0.05} & \textbf{75.10 $\pm$ 1.65} & \textbf{97.43 $\pm$ 0.08} \\
    \midrule
    VGG16 & SGDrop (curriculum $\rho$)&  81.82 $\pm$ 0.13 & 56.00 $\pm$ 0.22        & 78.80 $\pm$ 0.49 & 85.07 $\pm$ 0.42 & 98.65 $\pm$ 0.16 \\
    ResNet50 & SGDrop (curriculum $\rho$) &  77.14 $\pm$ 0.33 & 54.63  $\pm$ 0.60     & 72.61 $\pm$ 0.42 & 82.22 $\pm$ 0.71 & 98.58 $\pm$ 0.06 \\
    ConvNeXt & SGDrop (curriculum $\rho$) &  65.34 $\pm$ 0.46 & 29.69  $\pm$ 0.68     & 53.06 $\pm$ 0.36 & 75.49 $\pm$ 0.91 & 97.12 $\pm$ 0.04 \\
    \bottomrule
  \end{tabular}
  \end{adjustbox}
  \label{tab:Comparison_with_methods_across_datasets}
\end{table*}

\begin{table}[ht]
  \begin{minipage}[b]{0.48\linewidth}
    \caption{Accuracy on the testing sets (CIFAR10 and STL10).}
    \label{tab:cifar_stl10}
    \centering
    \begin{tabular}{@{}lcc@{}}
      \toprule
      Method             & CIFAR-10      & STL10 \\
      \midrule
      Vanilla           &   79.11 $\pm$ 0.24           & 63.24 $\pm$ 0.24 \\
      Dropout            &   \textbf{81.60 $\pm$ 0.18}  & 62.76 $\pm$ 0.88 \\
      Ex. Dropout        &   81.29 $\pm$ 0.17           & 63.38 $\pm$ 0.18 \\
      SGDrop  (ours)           &   81.57 $\pm$ 0.12           & \textbf{68.25 $\pm$ 0.48} \\
      \bottomrule
    \end{tabular}
  \end{minipage}
  \hfill
  \begin{minipage}[b]{0.48\linewidth}
    \caption{Transfer learning experiment RESISC45 $\rightarrow$ RSICD.}
    \label{tab:transfer_learning_exps_on_rsicd}
    \centering
    \begin{tabular}{@{}lcc@{}}
      \toprule
      Method & \small RESISC45 $\rightarrow$ RSICD \\
      \midrule
      Vanilla                   & 64.18 $\pm$ 0.50 \\
      Dropout                    & 65.69 $\pm$ 0.09 \\
      Ex. Dropout                & 64.64 $\pm$ 0.14 \\
      SGDrop   (ours)                  & \textbf{70.35 $\pm$ 0.37} \\
      \bottomrule
    \end{tabular}
  \end{minipage}
\end{table}

\begin{table}[ht!]
    \centering
    \caption{Comparison of VGG16 and ResNet50 generalization performance on ImageNet and Places365.}
    \begin{adjustbox}{width=\columnwidth,center}
    \begin{tabular}{@{}llcc|cc@{}}
    \toprule
    & & \multicolumn{2}{c|}{Without data augmentation} & \multicolumn{2}{c}{With data augmentation} \\
     Network    & Method & ImageNet & Places365 & ImageNet & Places365\\
    \midrule
      \multirow{3}{4em}{VGG16} & Vanilla & 39.46 $\pm$ 0.14 & 37.68 $\pm$ 0.25 & 63.60 $\pm$ 0.17 & 47.18 $\pm$ 0.22 \\
        & Dropout & 47.35 $\pm$ 0.10 & 42.98 $\pm$ 0.23 & 64.57 $\pm$ 0.17 & 47.94 $\pm$ 0.16 \\
        & SGDrop & \textbf{58.80 $\pm$ 0.48} & \textbf{46.35 $\pm$ 0.13} & \textbf{66.97 $\pm$ 0.30} & \textbf{54.62 $\pm$ 0.03} \\
    \midrule
    \midrule
        \multirow{3}{4em}{ResNet50} & Vanilla & 67.84 $\pm$ 0.16 & \textbf{46.81 $\pm$ 0.13} & 74.19 $\pm$ 0.24 & \textbf{54.73 $\pm$ 0.34} \\
        & Dropout & \textbf{68.70 $\pm$ 1.11} & 46.20 $\pm$ 0.08 & \textbf{74.00 $\pm$ 0.14} & 54.57 $\pm$ 0.15 \\
        & SGDrop & \textbf{68.70 $\pm$ 1.14} & 46.64 $\pm$ 0.17 & 73.93 $\pm$ 0.13 & 54.36 $\pm$ 0.10 \\
    \bottomrule
    \end{tabular}
    \end{adjustbox}
    \label{tab:large_data}
\end{table}

\subsection{Scheduling $\rho$}

It is noteworthy that SGDrop not only surpasses standard Dropout and Excitation Dropout in terms of generalization performance but also achieves this with a significantly reduced number of masked neurons. 
The adequate number of neurons to be dropped is guided by the hyper-parameter $\rho$, which establishes the quantile threshold for selective neuron deactivation.
Diverging from the random deactivation approach of traditional Dropout, SGDrop strategically targets over-specialized neurons using $\rho$, thereby enhancing the adaptability of the remaining ones.
Intuitively, higher $\rho$ values prompt the network to engage a wider variety of features, countering over-reliance on specific neurons, but also reducing the number of degrees of freedom for network optimization.
Our empirical findings, reported in Table \ref{tab:rho_study}, indicate that overly high $\rho$ values (exceeding 0.025 for VGG networks and 0.2 for ResNets) can impede learning.
To address this, we adopt a curriculum learning strategy, linearly increasing $\rho$ across epochs, from $\rho_{init}=0.01$ to $\rho_{final}=0.1$.
This method effectively mitigates the performance drop associated with high $\rho$ settings and further optimizes SGDrop’s impact.
The last line of Table \ref{tab:Comparison_with_methods_across_datasets} demonstrates that employing this curriculum approach with SGDrop leads to improvements in performance which are modest to significant compared to the baselines and to SGDrop with fixed small $\rho$. 
\begin{table*}
  \centering
  \caption{Influence of $\rho$ on SGDrop's generalization enhancement.}
  \begin{adjustbox}{width=\columnwidth,center}
  \begin{tabular}{@{}llcccccc@{}}
    \toprule
    Dataset & Model & $\rho=0.01$ & $\rho=0.025$ & $\rho=0.05$ & $\rho=0.075$ & $\rho=0.1$ & $\rho=0.2$\\
    \midrule
    \multirow{2}{5em}{RSICD} & VGG16     & 81.00 $\pm$ 0.44 & 78.62 $\pm$ 0.31 & 9.75 $\pm$ 0.00  & 9.75 $\pm$ 0.00 & 9.75 $\pm$ 0.00 & 9.75 $\pm$ 0.00 \\
    &ResNet50  & 77.54 $\pm$ 0.45 & 76.71 $\pm$ 0.59 & 77.51 $\pm$ 0.99 & 77.09 $\pm$0.46 & 76.39 $\pm$ 1.23 & 74.58 $\pm$1.79 \\      
    \midrule
    \multirow{2}{5em}{Imagenette} & VGG16     & 72.58 $\pm$ 1.55 & 72.37 $\pm$ 1.98 & 18.32 $\pm$ 24.56 & 6.04 $\pm$ 0.00 & 6.04 $\pm$ 0.00 & 6.04 $\pm$ 0.00 \\
    & ResNet50  & 72.61 $\pm$ 0.42 & 72.63 $\pm$ 0.01 & 73.38 $\pm$ 2.11 & 72.32 $\pm$ 0.69 & 72.78 $\pm$ 1.23 & 67.34 $\pm$ 0.09 \\ 
    \bottomrule
  \end{tabular}
  \end{adjustbox}
  \label{tab:rho_study}
\end{table*}

\subsection{Is the EMA useful?}

To enhance training stability, we employ an exponential moving average (EMA) model for computing the dropout masks in SGDrop.
Table \ref{tab:ablation_study_combined} reports the generalization performance difference on the Imagenette and RSICD datasets, between computing the SGDrop masks from the trained model $\psi$ or from the EMA model $\psi'$. 
Notably, even without EMA, SGDrop significantly outperforms the baseline for the VGG16 and ConvNext architectures, achieving respectively a remarkable improvement of $+13.31\%$ and $+8,07\%$ on Imagenette and $+16.74\%$ and $+5.44\%$on RSICD.
For the ResNet50 architecture, SGDrop shows a modest enhancement on Imagenette ($+0.29\%$) but a slight decrease on RSICD ($-1.55\%$).
However, the incorporation of EMA consistently enhances performance across both networks and datasets.
This demonstrates the efficacy of using EMA in SGDrop, highlighting its contribution to the method's overall effectiveness.
\begin{table}
  \centering
  \caption{Ablation experiment results on Imagenette and RSICD.}
  \begin{adjustbox}{width=\columnwidth,center}
  \begin{tabular}{@{}lccc|ccc@{}}
    \toprule
    Dataset & \multicolumn{3}{c|}{Imagenette} & \multicolumn{3}{c}{RSICD} \\
    \midrule
    Variant & VGG16 & ResNet50  & ConvNeXt & VGG16 & ResNet50 & ConvNeXt\\
    \midrule
    Vanilla               & 67.21 $\pm$ 0.29             & 74.96 $\pm$ 0.46          & 55.08 $\pm$ 2.01   & 55.60 $\pm$ 2.13  & 69.35 $\pm$ 1.43         & 49.04 $\pm$ 1.01 \\
    SGDrop w/o EMA        & 80.52 $\pm$ 0.05             & 75.25 $\pm$ 0.41          & 63.85 $\pm$ 2.15  & 72.34 $\pm$ 0.69   & 67.80 $\pm$ 1.31         & 54.48 $\pm$ 1.00\\
    SGDrop with EMA      & \textbf{81.00 $\pm$ 0.44}    & \textbf{77.54 $\pm$ 0.45} & \textbf{67.42 $\pm$ 2.12}  & \textbf{75.94 $\pm$ 2.28} & \textbf{71.06 $\pm$ 1.11} & \textbf{55.76 ± 0.05}\\
     \bottomrule
  \end{tabular}
  \end{adjustbox}
  \label{tab:ablation_study_combined}
\end{table}

\subsection{Computation overhead}

Compared to vanilla training, or even Dropout, SGDrop necessitates computing the attribution mask and applying it to the feature map, before we obtain a loss value and can backpropagate and apply gradients on network weights. 
Computing the attribution mask itself requires expanding the computation graph with nodes that actually correspond to a partial backpropagation from $\psi_c$'s output (gradients computed but not applied), in order to obtain $\nabla_z \psi_c$. 
This (and to a much lower extent, the EMA) can be expected to induce some computational overhead. 
To assess the overall impact of this extra computation, Table \ref{tab:overhead} reports the training time required per batch of 64 images on VGG16, ResNet50 and ConvneXt architectures, on the same computing resource, for vanilla training and SGDrop.
SGDrop appears to consistently induce a computation overhead of about $33\%$ across contexts, which may be deemed acceptable given the increase in generalization performance.

\begin{table}[!ht]
  \centering
  \caption{SGDrop computational overhead.}
  \begin{tabular}{@{}lccc@{}}
    \toprule
    Architecture & Vanilla & SGDrop & Overhead \\
    \midrule
    VGG16    & 281.43 ms & 375.00 ms & +33.24\%  \\
    ResNet50 & 155.09 ms & 207.50 ms & +30.44\%  \\
    ConvNeXt & 438.30 ms & 548.20 ms & +25.07\%  \\
    \bottomrule
  \end{tabular}
  \label{tab:overhead}
\end{table}

\section{Limitations and perspectives}    
\label{sec:limitations}
While SGDrop enables considerable improvements in network interpretability and generalization, particularly in architectures like VGG16 or ConvNext, its limitations warrant attention.
Firstly, the method's effectiveness varies across different architectures, with less pronounced benefits in models like ResNet50. 
Conversely, the fact that SGDrop enables VGG16 networks to outperform ResNet50 ones also begs for further analysis of how architectural patterns may actually prevent network optimization from achieving good generalization scores. 
Additionally, the choice of the hyper-parameter $\rho$ is crucial and requires careful tuning, which may not be straightforward across varying datasets and tasks.
The computational overhead, primarily due to the calculation of attribution maps, is another concern, potentially affecting training efficiency in large-scale models or datasets.
Lastly, our implementation currently focuses on image classification tasks, leaving its efficacy in other domains or complex tasks such as object detection or segmentation unexplored.
It may be questioned why we did not implement SGDrop in transformer models, which represent the state-of-the-art in computer vision. Firstly, it is important to highlight that, in scenarios with limited data, transformers tend to underperform compared to CNNs \cite{d2021convit, park2021vision, raghu2021vision, hassani2021escaping, liu2021efficient, lu2022bridging}, as detailed in the Appendix. Secondly, our initial experiments with a Vision Transformer (ViT) did not exhibit the same degree of over-concentration on specific image regions accross trainnig eopchs that was observed with CNNs, suggesting that the benefits of SGDrop may not translate as effectively to transformer architectures.
In particular, SGDrop's effectiveness is contingent upon attribution methods that pinpoint critical regions within the feature space. Research has demonstrated these methods to be less effective for transformers \cite{zhang2023opti}. This implies that the dropout selection mechanism of SGDrop, which excels with CNNs, might require re-evaluation for transformers. 
Research has demonstrated that these methods are less effective for transformers, which suggests that the dropout mechanism employed by SGDrop may need reevaluation for such architectures.
A promising alternative could involve utilizing attention matrices rather than attribution maps, following the approach outlined by Yang et al. \cite{yang2022ad}. 
\section{Conclusion}
\label{sec:conclusion}
In this work, we have demonstrated that convolutional neural networks tend to narrow their focus on minute details over the course of training iterations.
This phenomenon, often leading to over-specialization, is traditionally mitigated by methods like Dropout which myopically deactivate features at random during learning, to force dispatching of feature influence.
In contrast, our saliency-guided dropout (SGDrop) method adopts a more focused approach by selectively targeting the most influential neurons. 
Crucially, this enables the remaining neurons to adapt and evolve, fostering a more versatile and robust network.
Our findings reveal significant improvements in the attributions generated by SGDrop. These attributions are not only wider but also align more closely with labels and human interpretations, as evidenced by our comprehensive analysis.
The practical outcome of these advancements is a notable improvement in model generalization. 
Across various datasets and network architectures, SGDrop consistently outperforms standard approaches, underlining its effectiveness in diverse scenarios.
Moreover, SGDrop's versatility lies in its generic nature.
It offers ample scope for adaptation and refinement, be it in attribution computation methods, feature levels selected for deactivation, or the strategy for neuron dropping.
This flexibility paves the way for future research to explore and expand upon SGDrop's foundation, potentially leading to even more robust and adaptable neural network models.

\section*{Acknowledgments}
The authors acknowledge the support of the DEEL, and MINDS projects, the funding of the AI Interdisciplinary Institute ANITI
funding, through the French “Investing for the Future – PIA3” program under grant agreement
ANR-19-PI3A-0004.
This work benefited from computing resources from CALMIP under grant P21001.

\clearpage  

%
%
\bibliographystyle{splncs04}
\bibliography{main}
\clearpage 
\appendix
\section{Transformers generalization performance in low data-regime}
\label{sec:transformers}
To corroborate the assertion that CNNs outperform transformer-based models in low data-regime scenarios, we assessed the generalization performance of various Vision Transformer (ViT) \cite{dosovitskiy2020image} architectures in contrast to convolutional neural network models employed in the experimental sections of this paper. The results, as depicted in Table \ref{tab:transformers}, demonstrate that CNNs, particularly VGG16 and ResNet50, exhibit superior generalization in low data regimes relative to their ViT counterparts.

\begin{table*}[!ht]
  \centering
  \caption{Accuracy on the testing sets.}
  \small
  \begin{tabular}{@{}llccccc@{}}
    \toprule
    Network                        & Imagenette                & Imagewoof              & RSICD              & RESISC45           & PatternNet       \\
    \midrule
    VGG16   & 67.21 $\pm$ 0.29          &  43.65 $\pm$ 0.67      & 55.60 $\pm$ 2.13   & 59.32 $\pm$ 2.71   & 90.35 $\pm$ 1.37  \\
    ResNet50      & 74.96 $\pm$ 0.46          &  45.76 $\pm$ 0.13      & 69.35 $\pm$ 1.43   & 78.28 $\pm$ 0.48   & 98.52 $\pm$ 0.21  \\
    ConvNeXt & 55.08 $\pm$ 2.01          &  24.64 $\pm$ 0.49      & 49.04 $\pm$ 1.01   & 67.89 $\pm$ 0.88   & 95.96 $\pm$ 0.27  \\
    \midrule
    ViT B 16                       & 61.11 $\pm$ 0.62          &  34.58 $\pm$ 1.40      & 61.34 $\pm$ 0.23   & 73.47 $\pm$ 0.34   & 94.32 $\pm$ 0.35  \\
    ViT B 32                       & 54.81 $\pm$ 0.99          &  29.83 $\pm$ 0.27      & 54.07 $\pm$ 0.82   & 64.52 $\pm$ 0.67   & 90.92 $\pm$ 0.17  \\
    ViT L 16                       & 57.95 $\pm$ 0.19                &  32.33 $\pm$ 1.55             & 54.89 $\pm$ 0.91   & 74.44 $\pm$ 0.41   & 91.91 $\pm$ 0.25  \\
    ViT L 32                       & 53.77 $\pm$ 0.34          &  29.00 $\pm$ 0.49      & 52.20 $\pm$ 0.50   & 65.36 $\pm$ 0.94   & 90.14 $\pm$ 0.12  \\
    \bottomrule
  \end{tabular}
  \caption{Comparison of ViT and convolutional networks accuracy on the testing sets.}
  \label{tab:transformers}
\end{table*}

\section{Experimental Setting}
\label{sec:experimental_settings}


\begin{table}[h!t]
\centering
\small
\begin{tabular}{|l|l|l|l|l|l|}
\hline
\textbf{Dataset} & \textbf{Model} & \textbf{Epochs} & \textbf{Optimizer} & \textbf{Learning Rate} & \textbf{\(\rho\)} \\ \hline
CIFAR & cifarmodel & 100 & Adam & lr=0.0001 & 0.01\\ \hline
STL10 & VGG9 & 80 & Adam & lr=0.00005 & 0.01\\ \hline
\multirow{2}{*}{Imagenette} & ResNet50/VGG16 & 80 & Adam & lr=0.0001 & 0.01\\ \cline{2-6}
& convnext & 80 & Adam & lr=0.0001 & 0.1\\ \hline
\multirow{2}{*}{Imagewoof} & ResNet50/VGG16 & 80 & Adam & lr=0.0001 & 0.01\\ \cline{2-6}
& convnext & 80 & Adam & lr=0.0001 & 0.1\\ \hline
\multirow{3}{*}{RSICD} & ResNet50/VGG16 & 80 & Adam & lr=0.0001 & 0.01\\ \cline{2-6}
& \multirow{2}{*}{convnext} & \multirow{2}{*}{80} & \multirow{2}{*}{Adam} & \multirow{2}{*}{lr=0.0001} & \multirow{2}{*}{0.05}\\ 
& & & & & \\ \hline
\multirow{3}{*}{RESISC45} & ResNet50/VGG16 & 80 & Adam & lr=0.0001 & 0.01\\ \cline{2-6}
& \multirow{2}{*}{convnext} & \multirow{2}{*}{80} & \multirow{2}{*}{Adam} & \multirow{2}{*}{lr=0.0001} & \multirow{2}{*}{0.05}\\
& & & & & \\ \hline
\multirow{3}{*}{PatternNet} & ResNet50/VGG16 & 80 & Adam & lr=0.0001 & 0.01\\ \cline{2-6}
& \multirow{2}{*}{convnext} & \multirow{2}{*}{80} & \multirow{2}{*}{Adam} & \multirow{2}{*}{lr=0.0001} & \multirow{2}{*}{0.05}\\
& & & & & \\ \hline
ImageNet & ResNet50/VGG16 & 90 & SGD & \begin{tabular}[c]{@{}l@{}}lr=0.1/0.01/0.01\end{tabular}  & 0.01\\ \hline
Places365 & ResNet50/VGG16 & 90 & SGD & \begin{tabular}[c]{@{}l@{}}lr=0.1/0.01/0.01\end{tabular} & 0.01\\ \hline
\end{tabular}
\caption{Experimental Settings}
\label{tab:experimental_settings}
\end{table}

Table \ref{tab:experimental_settings} presents a detailed summary of the experimental settings employed in Section 4. In our CIFAR dataset experiments, we utilized the network architecture as described in \cite{zunino2021excitation}. This architecture encompasses three convolutional blocks, each consisting of convolutional filters with 96, 128, and 256 filters of size 5×55×5, a stride of 1, and padding of 2. Each block is followed by a 3×33×3 max-pooling layer with a stride of 2. The network concludes with two fully-connected layers, each with 2048 units, and a softmax output layer comprising 10 units.
For the ImageNet and Places365 experiments, we used the official PyTorch script for ImageNet training which includes a dynamic learning rate schedule, which entails dividing the learning rate by 10 after every 30 epoch.

\end{document}